\documentclass[10pt,twocolumn,letterpaper]{article}

\usepackage{cvpr}
\usepackage{times}
\usepackage{epsfig}
\usepackage{graphicx}
\usepackage{amsmath}
\usepackage{amssymb}

\usepackage{lipsum}
\usepackage{comment}
\usepackage{xcolor}
\usepackage{adjustbox}
\usepackage{multirow}
\usepackage[labelsep=period]{caption}
\usepackage{subcaption}
\usepackage{makecell}
\usepackage{romannum}

\usepackage{pdfpages}

\DeclareMathAlphabet\mathbfcal{OMS}{cmsy}{b}{n}

\usepackage[pagebackref=true,breaklinks=true,letterpaper=true,colorlinks,bookmarks=false]{hyperref}

\cvprfinalcopy 

\def\cvprPaperID{8918} 

\ifcvprfinal\fi
\begin{document}
\pagenumbering{arabic}

\title{Embedding Expansion: Augmentation in Embedding Space \\ for Deep Metric Learning}

\author{Byungsoo Ko\thanks{Authors contributed equally.},\space\space Geonmo Gu\footnotemark[1]\\
NAVER/LINE Vision\\
{\tt\small \{kobiso62, korgm403\}@gmail.com}
}

\maketitle

\begin{abstract}
Learning the distance metric between pairs of samples has been studied for image retrieval and clustering.
With the remarkable success of pair-based metric learning losses~\cite{chopra2005contrastive, weinberger2009triplet, sohn2017distance, oh2016deep}, recent works~\cite{zheng2019hardness, duan2018deep, zhao2018adversarial} have proposed the use of generated synthetic points on metric learning losses for augmentation and generalization.
However, these methods require additional generative networks along with the main network, which can lead to a larger model size, slower training speed, and harder optimization.
Meanwhile, post-processing techniques, such as query expansion~\cite{chum2007QE, chum2011QE} and database augmentation~\cite{turcot2009DBA, arandjelovic2012DBA}, have proposed a combination of feature points to obtain additional semantic information.
In this paper, inspired by query expansion and database augmentation, we propose an augmentation method in an embedding space for metric learning losses, called \textbf{embedding expansion}.
The proposed method generates synthetic points containing augmented information by a combination of feature points and performs hard negative pair mining to learn with the most informative feature representations.
Because of its simplicity and flexibility, it can be used for existing metric learning losses without affecting model size, training speed, or optimization difficulty.
Finally, the combination of embedding expansion and representative metric learning losses outperforms the state-of-the-art losses and previous sample generation methods in both image retrieval and clustering tasks.
The implementation is publicly available\footnote{\url{https://github.com/clovaai/embedding-expansion}}.

\end{abstract}

\section{Introduction}

\begin{figure}[t]
\centering
\includegraphics[width=0.88\columnwidth]{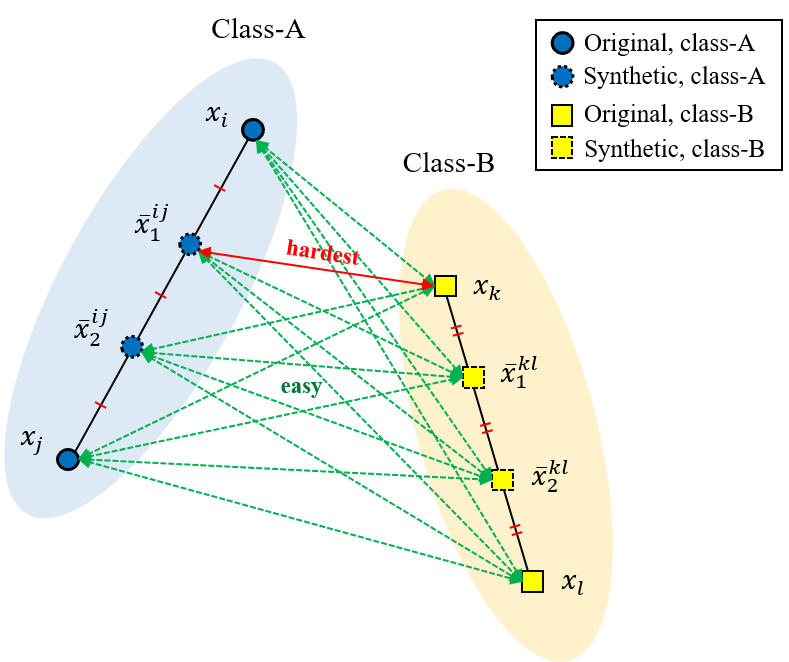}
\caption{Illustration of our proposed embedding expansion consisting of two steps.
In the first step, given a pair of embedding points from the same class, we perform linear interpolation on the line of the embedding points to generate internally dividing synthetic points into $n+1$ equal parts, where $n$ is the number of synthetic points ($n=2$ in the Figure).
Secondly, we select the hardest negative pair within the possible negative pairs of original and synthetic points.
Rectangles and circles represent the two different classes, where the plain boundary indicates original points ($x_i, x_j, x_k, x_l$), and the dotted boundary indicates synthetic points ($\bar{x}^{ij}_1, \bar{x}^{ij}_2, \bar{x}^{kl}_1, \bar{x}^{kl}_2$).}
\label{fig:teaser}
\end{figure}

Deep metric learning aims to learn a distance metric for measuring similarities between given data points.
It has played an important role in a variety of applications in computer vision, such as image retrieval~\cite{wang2014learning, gordo2016deep}, re-identification~\cite{zheng2015scalable, li2014deepreid, yi2014deep}, clustering~\cite{hershey2016deep}, and face recognition~\cite{chopra2005contrastive, schroff2015facenet, taigman2014deepface}.
The core idea of deep metric learning is to learn an embedding space by pulling the same class samples together and by pushing different class samples apart.
To learn an embedding space, many of the metric learning losses take pairs of samples to optimize the loss with the desired properties.
Conventional pair-based metric learning losses are contrastive loss~\cite{chopra2005contrastive, hadsell2006dimensionality} and triplet loss~\cite{weinberger2009triplet, schroff2015facenet}, which take 2-tuple and 3-tuple samples, respectively.
N-pair loss~\cite{sohn2017distance} and lifted structured loss~\cite{oh2016deep} aim to exploit a greater number of negative samples to improve the conventional metric learning losses.
Recent works~\cite{wang2017deep, ustinova2016learning, wu2017sampling, oh2017deep} have been proposed to consider the richer structured information among multiple samples.

Along with the importance of the loss function, the sampling strategy also plays an important role in performance.
Different strategies for the same loss function can lead to extremely different results~\cite{wu2017sampling, yu2018hard}.
Thus, there has been active research on sampling strategy and hard sample mining methods~\cite{wu2017sampling, ahmed2015improved, hermans2017defense, schroff2015facenet}.
One drawback of sampling and mining strategies is that it can lead to a biased model due to training with a minority of selected hard samples and ignoring a majority of easy samples~\cite{wu2017sampling, schroff2015facenet, zheng2019hardness}.
To address this problem, hard sample generation methods~\cite{zheng2019hardness, duan2018deep, zhao2018adversarial} have been proposed to generate hard synthesis with easy samples.
However, those methods require an additional sub-network as a generator, such as a generative adversarial network and an auto-encoder, which can cause a larger model size, slower training speed, and more training difficulty~\cite{arjovsky2017wasserstein}. 

In this paper, we propose a novel augmentation method in the embedding space for deep metric learning, called embedding expansion (EE).
Inspired by query expansion~\cite{chum2007QE, chum2011QE} and database augmentation techniques~\cite{turcot2009DBA, arandjelovic2012DBA}, the proposed method combines feature points to generate synthetic points with augmented image representations.
As illustrated in Figure~\ref{fig:teaser}, it generates internally dividing points into $n+1$ equal parts within pairs of the same classes and performs hard negative pair mining among original and synthetic points.
By exploiting synthetic points with augmented information, it attains a performance boost through a more generalized model.
The proposed method is simple and flexible enough that it can be combined with existing pair-based metric learning losses.
Unlike the previous sample generation method, the proposed method does not suffer from the problems caused by using an additional generative network, because it performs simple linear interpolation for sample generation.
We demonstrate that combining the proposed method with existing metric learning losses achieves a significant performance boost, while it also outperforms the previous sample generation methods on three famous benchmarks (CUB200-2011~\cite{wah2011caltech}, CARS196~\cite{krause20133d}, and Stanford Online Products~\cite{oh2016deep}) in both image retrieval and clustering tasks.

\section{Related Work}

\paragraph{Sample Generation}
Recently, there have been attempts to generate potential hard samples for pair-based metric learning losses~\cite{zheng2019hardness, duan2018deep, zhao2018adversarial}.
The main purpose of generating samples is to exploit a large number of easy negatives and train the network with this extra semantic information.
The deep adversarial metric learning (DAML) framework~\cite{duan2018deep} and the hard triplet generation (HTG)~\cite{zhao2018adversarial} use generative adversarial networks to generate synthetic samples.
The hardness-aware deep metric learning (HDML) framework~\cite{zheng2019hardness} exploits an auto-encoder to generate label-preserving synthesis and control the hardness of synthetic negatives. 
Even though training with synthetic samples generated by the above methods can give a performance boost, they require additional generative networks alongside the main network.
This can result in a larger model size, slower training time, and harder optimization~\cite{arjovsky2017wasserstein}.
The proposed method also generates samples to train with augmented information, while it does not require any additional generative networks and suffer from the above problems.

\paragraph{Query Expansion and Database Augmentation}

Query expansion (QE) in image retrieval has been proposed in~\cite{chum2007QE, chum2011QE}.
Given a query image feature, it retrieves a rank list of image features from a database that matches the query and combines the high ranked retrieved image features, along with the original query.
Then, it re-queries the combined image features to retrieve an expanded set of matching images and repeats the process as necessary.
Similar to query expansion, database augmentation (DBA)~\cite{turcot2009DBA, arandjelovic2012DBA} replaces every image feature in a database with a combination of itself and its neighbors, to improve the quality of image features.
Our proposed embedding expansion is inspired by these concepts namely the combination of image features to augment image representations by leveraging the features of their neighbors.
The key difference is that both techniques are used during post-processing, while the proposed method is used during the training phase.
More specifically, the proposed method generates multiple combinations from the same class to augment semantic information for metric learning losses.

\section{Preliminaries}

This section introduces the mathematical formulation of the representative pair-based metric learning losses.
We define a function $f$ which projects data space $\mathbfcal{D}$ to the embedding space $\mathbfcal{X}$ by $f(\cdot;\theta) : \mathbfcal{D} \xrightarrow{} \mathbfcal{X}$, where $f$ is a neural network parameterized by $\theta$.
Feature points in the embedding space can be sampled as $\mathbf{X} = [\mathbf{x}_1, \mathbf{x}_2, \dots, \mathbf{x}_N]$, where $N$ is the number of feature points and each point $\mathbf{x}_i$ has a label $\mathbf{y}[i] \in \{1, \dots, C\}$.
$\mathcal{P}$ is a set of positive pairs among the feature points.

\textbf{\textit{Triplet loss}}~\cite{weinberger2009triplet, schroff2015facenet} considers triplet of points and pulls the anchor point closer to the positive point of the same class than to the negative point of the different class by a fixed margin $m$:
\begin{align}
L_{triplet} = \frac{1}{|\mathcal{P}|}\sum_{\substack{(i, j)\in{\mathcal{P}} \\ k:\mathbf{y}[k] \neq \mathbf{y}[i]}}\Big[ d_{i,j}^2 - d_{i,k}^2 + m \Big]_+,
\end{align}
where $[\cdot]_+$ is a hinge function and $d_{i,j} = \| \mathbf{x}_i - \mathbf{x}_j \|_2$ is the Euclidean distance between embedding $\mathbf{x}_i$ and $\mathbf{x}_j$.
Triplet loss is usually used with applying $L_2$-normalization to the embedding feature~\cite{schroff2015facenet}.

\textbf{\textit{Lifted structured loss}}~\cite{oh2016deep} is proposed to take full advantage of the training batches in the neural network training.
Given a training batch, it aims to pull one positive point as close as possible and pushes all negative points corresponding to the positive points farther than a fixed margin of $m$:
\begin{align}
L_{lifted} & = \frac{1}{2|\mathcal{P}|} \sum_{(i,j) \in \mathcal{P}}\bigg[\log\bigg\{ \sum_{k:\mathbf{y}[k] \neq \mathbf{y}[i]}exp{\big(m-d_{i,k}\big)} \nonumber \\
& + \sum_{k:\mathbf{y}[k] \neq \mathbf{y}[j]}exp{\big(m - d_{j,k} \big)}\bigg\} + d_{i,j} \bigg]^2_+.
\end{align}
Similar to triplet loss, lifted structured loss also uses $L_2$-normalization to the embedding feature~\cite{oh2016deep}.

\textbf{\textit{N-pair loss}}~\cite{sohn2017distance} allows joint comparison among more than one negative points to generalize triplet loss.
More specifically, it aims to pull one positive pair and push away $N-1$ negative points from $N-1$ negative classes:
\begin{align}
L_{npair} = \frac{1}{|\mathcal{P}|} \sum_{(i,j) \in \mathcal{P}}\bigg\{ \log\bigg[ 1 + \sum_{k:\mathbf{y}[k] \neq \mathbf{y}[i]}exp{\big(s_{i,k} - s_{i,j} \big)}\bigg] \bigg\},
\end{align}
where $s_{i,j}={\mathbf{x}_i}^T \mathbf{x}_j$ is the similarity of embedding points $\mathbf{x}_i$ and $\mathbf{x}_j$.
N-pair loss does not apply $L_2$-normalization to the embedding features because it leads to optimization difficulty for the loss~\cite{sohn2017distance}.
Instead, it regularizes the $L_2$ norm of the embedding features to be small.

\textbf{\textit{Multi-Similarity loss}}~\cite{wang2019multi} (MS loss) is one of the latest works for metric learning loss.
It is proposed to jointly measure both self-similarity and relative similarities of a pair, which enables the model to collect and weight informative pairs.
MS loss performs pair mining for both positive and negative pairs.
A negative pair of $\{\mathbf{x}_i, \mathbf{x}_j\}$ is selected with the condition of 
\begin{align}
s_{i,j}^{-} > \min_{\mathbf{y}[k]=\mathbf{y}[i]}s_{i,k}-\epsilon,
\label{eq:ms_neg_mining}
\end{align}
and a positive pair of $\{\mathbf{x}_i, \mathbf{x}_j\}$ is selected with the condition of 
\begin{align}
s_{i,j}^{+} < \max_{\mathbf{y}[k] \neq \mathbf{y}[i]}s_{i,k}+\epsilon,
\end{align}
where the $\epsilon$ is a given margin.
For an anchor $x_i$, we denote the index set of its selected positive and negative pairs as $\widetilde{\mathcal{P}_i}$ and $\widetilde{\mathcal{N}_i}$, respectively.
Then, MS loss can be formulated as:
\begin{align}
L_{ms} & = \frac{1}{N} \sum_{i=1}^{N}\bigg[\frac{1}{\alpha}\log\bigg\{ 1 + \sum_{k \in \widetilde{\mathcal{P}_i}}e^{-\alpha\big(s_{i,k}-\lambda\big)}\bigg\} \nonumber \\
& + \frac{1}{\beta}\log\bigg\{\sum_{k \in \widetilde{\mathcal{N}_i}}e^{\beta\big(s_{i,k}-\lambda\big)}\bigg\}\bigg],
\end{align}
where $\alpha$, $\beta$, and $\lambda$ are hyper-parameters, and $N$ denotes the number of training samples.
MS loss uses $L_2$-normalization on the embedding features.

\section{Embedding Expansion}
This section introduces the proposed embedding expansion consisting of two steps: synthetic points generation and hard negative pair mining.

\subsection{Synthetic Points Generation}

\begin{figure}[t]
\centering
\includegraphics[width=0.75\columnwidth]{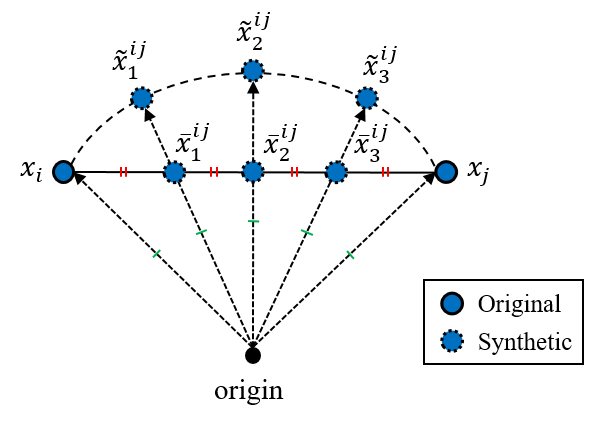} 
\caption{Illustration of generating synthetic points. Given two feature points $\{x_i, x_j\}$ from the same class, embedding expansion generates synthetic points which are internally dividing points into $n+1$ equal parts $\{\bar{x}^{ij}_1, \bar{x}^{ij}_2, \bar{x}^{ij}_3\}$, where $n=3$ in the figure. For the metric learning losses that use $L_2$-normalization, the synthetic points are applied with $L_2$-normalization and generates $\{\Tilde{x}^{ij}_1, \Tilde{x}^{ij}_2, \Tilde{x}^{ij}_3\}$. Circles with plane line are original points and circles with dotted line are synthetic points.}
\label{fig:generation}
\end{figure}

QE and DBA techniques in image retrieval generate synthetic points by combining feature points in an embedding space in order to exploit additional relevant information~\cite{chum2007QE, chum2011QE, turcot2009DBA, arandjelovic2012DBA}.
Inspired by these techniques, the proposed embedding expansion generates multiple synthetic points by combining feature points from the same class in an embedding space to augment information for the metric learning losses.
To be specific, embedding expansion performs linear interpolation in a linear interpolant between two feature points and generates synthetic points that are internally dividing points into $n+1$ equal parts, as illustrated in Figure~\ref{fig:generation}.

Given two feature points $\{\mathbf{x}_i, \mathbf{x}_j\}$ from the same class in an embedding space, the proposed method generates internally dividing points $\bar{\mathbf{x}}_k^{ij}$ into $n+1$ equal parts and obtains a set of the synthetic points $\mathcal{\bar{S}}^{ij}$ as:
\begin{align}
\bar{\mathbf{x}}_k^{ij}=\frac{k\mathbf{x}_i+(n-k)\mathbf{x}_j}{n}, \\
\mathcal{\bar{S}}^{ij}=\{\bar{\mathbf{x}}_1^{ij}, \bar{\mathbf{x}}_2^{ij}, \cdots \bar{\mathbf{x}}_n^{ij} \},
\end{align}
where $n$ is the number of points to generate.
For the metric learning losses that use $L_2$-normalization, such as triplet loss, lifted structured loss, and MS loss, $L_2$-normalization has to be applied to the synthetic points:
\begin{align}
\tilde{\mathbf{x}}_k^{ij}&=\frac{\bar{\mathbf{x}}_k^{ij}}{\| \bar{\mathbf{x}}_k^{ij} \|_2}, \\
\mathcal{\widetilde{S}}^{ij}&=\{\tilde{\mathbf{x}}_1^{ij}, \tilde{\mathbf{x}}_2^{ij}, \cdots \tilde{\mathbf{x}}_n^{ij} \},
\end{align}
where $\tilde{\mathbf{x}}_k^{ij}$ is a $L_2$-normalized synthetic point, and $\mathcal{\widetilde{S}}^{ij}$ is a set of the $L_2$-normalized synthetic points.
These $L_2$-normalized synthetic points will be located on the hyper-sphere space with the same norm.
The way of generating synthetic points shares a similar spirit with mixup augmentation methods~\cite{zhang2017mixup, tokozume2018between, verma2018manifold}, and the comparison is given in supplementary material.

There are three advantages of generating points that are internally dividing points into $n+1$ equal parts in an embedding space.
(\romannum{1}) Given a pair of feature points from each class in well-clustered embedding space, the similarity of the hardest negative pair will be the shortest distance between line segments of each pair from each class (i.e., $\overline{\mathbf{x}_i\mathbf{x}_j} \leftrightarrow  \overline{\mathbf{x}_k\mathbf{x}_l}$ in Figure~\ref{fig:teaser}).
However, it is computationally expensive to compute the shortest distance between segments of finite length in a high-dimensional space~\cite{lumelsky1985fast, smith1916practical}.
Instead, by computing distances between internally dividing points of each class, we can approximate the problem with less computation.
(\romannum{2}) The labels of synthetic points have a high degree of certainty because they are included inside the class cluster.
Previous work~\cite{zheng2019hardness} of sample generation method exploited a fully connected layer and softmax loss to control the labels of synthetic points, while the proposed method makes it certain by considering geometrical relations.
We further investigate the certainty of labels of synthetic points with an experiment in Section~\ref{sec:label}.
(\romannum{3}) The proposed method of generating synthetic points requires a trivial amount of training speed and memory because we perform a simple linear interpolation in an embedding space.
We further discuss the training speed and memory in Section~\ref{sec:speedmemory}.

\subsection{Hard Negative Pair Mining}

The second step of the proposed method is to perform hard negative pair mining among the synthetic and original points to ignore trivial pairs and train with informative pairs, as illustrated in Figure~\ref{fig:teaser}.
The hard pair mining is only performed on negative pairs, and original points are used for positive pairs.
The reason is that hard positive pair mining among original and synthetic points will always be a pair of original points because the synthetic points are internally dividing points of the pair.
We formulate the combination of representative metric learning losses with the proposed embedding expansion.

\textbf{\textit{EE + Triplet loss}}~\cite{weinberger2009triplet, schroff2015facenet} can be formulated by adding min-pooling on the negative pairs because the hardest pair for triplet loss is a pair with the smallest Euclidean distance:
\begin{align}
L_{triplet}^{EE} = \frac{1}{|\mathcal{P}|}\hspace*{-0.1cm}\sum_{\substack{(i, j)\in\mathcal{P} \\ k:\mathbf{y}[k] \neq \mathbf{y}[i]}}\hspace*{-0.15cm}\Big[ d_{i,j}^2 - \hspace*{-0.2cm}\min_{\substack{(p, n)\in{\widehat{\mathcal{N}}_{\mathbf{y}[i],\mathbf{y}[k]}}}} \hspace*{-0.2cm} d_{p,n}^2 + m \Big]_+,
\end{align}
where $\widehat{\mathcal{N}}_{\mathbf{y}[i],\mathbf{y}[k]}$ is a set of negative pairs with a positive point from the class $\mathbf{y}[i]$ and a negative point from the class $\mathbf{y}[k]$ including synthetic points.

\textbf{\textit{EE + Lifted structured loss}}~\cite{oh2016deep} also has to use min-pooling of Euclidean distance of negative pairs to add embedding expansion.
The combined loss consists of minimizing the following hinge loss,
\begin{flalign}
&L_{lifted}^{EE} = \frac{1}{|\mathcal{P}|} \sum_{(i,j) \in \mathcal{P}}\bigg[\log\bigg\{ \nonumber \\
& \hspace*{-0.5cm}\sum_{k:\mathbf{y}[k] \neq \mathbf{y}[i]}exp{\big(m-\min_{(p,n)\in\widehat{\mathcal{N}}_{\mathbf{y}[i],\mathbf{y}[k]}}{d_{p,n}}\big)}\bigg\} + d_{i,j} \bigg]^2_+.
\end{flalign}

\textbf{\textit{EE + N-pair loss}}~\cite{sohn2017distance} can be formulated by using max-pooling on the negative pairs because the hardest pair for n-pair loss is a pair with the largest similarity, unlike triplet and lifted structured loss:
\begin{flalign}
&L_{npair}^{EE} = \frac{1}{|\mathcal{P}|} \sum_{(i,j) \in \mathcal{P}}\bigg\{ \log\bigg[ 1 \nonumber \\
& + \sum_{k:\mathbf{y}[k] \neq \mathbf{y}[i]}exp{\big(\max_{(p,n)\in\widehat{\mathcal{N}}_{\mathbf{y}[i],\mathbf{y}[k]}}s_{p,n} - s_{i,j} \big)}\bigg] \bigg\}.
\end{flalign}

\textbf{\textit{EE + Multi-Similarity loss}}~\cite{wang2019multi} contains two kinds of hard negative pair mining: one from the embedding expansion, and the other one from the MS loss.
We integrate both hard negative pair mining by modifying the condition of Equation~\ref{eq:ms_neg_mining}.
A negative pair of $\{\mathbf{x}_i, \mathbf{x}_j\}$ is selected with the condition of 
\begin{align}
\max_{(p,n)\in\widehat{\mathcal{N}}_{\mathbf{y}[i],\mathbf{y}[j]}}s_{p,n} > \min_{\mathbf{y}[k]=\mathbf{y}[i]}s_{i,k}-\epsilon,
\end{align}
and we define the index set of selected negative pairs of an anchor $x_i$ as $\widetilde{\mathcal{N}^{'}_i}$.
Then, the combination of embedding expansion and MS loss can be formulated as:
\begin{align}
L_{ms}^{EE} & = \frac{1}{N} \sum_{i=1}^{N}\bigg[\frac{1}{\alpha}\log\bigg\{ 1 + \sum_{k \in \widetilde{\mathcal{P}_i}}e^{-\alpha\big(s_{i,k}-\lambda\big)}\bigg\} \nonumber \\
& + \frac{1}{\beta}\log\bigg\{\sum_{k \in \widetilde{\mathcal{N}^{'}_i}}e^{\beta\big(s_{i,k}-\lambda\big)}\bigg\}\bigg].
\end{align}


\section{Experiments}

\begin{figure}[t!h!]
\centering
\includegraphics[width=0.8\columnwidth]{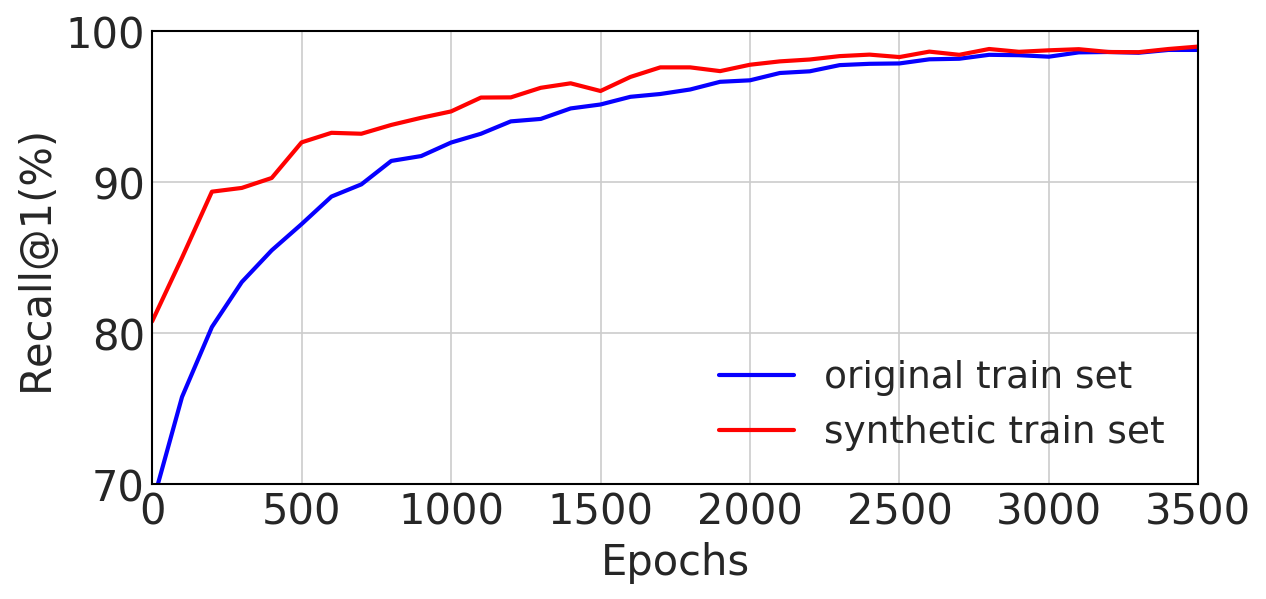} 
\caption{Recall@1(\%) curve evaluated with original and synthetic points from the train set, trained with EE + triplet loss on CARS196.}
\label{fig:label_synthetic}
\end{figure}

\subsection{Datasets and Settings} \label{sec:dataset}
\paragraph{Datasets}
We evaluate the proposed method with two small benchmark datasets (CUB200-2011~\cite{wah2011caltech}, CARS196~\cite{krause20133d}), and one large benchmark dataset (Stanford Online Products~\cite{oh2016deep}).
We follow the conventional way of train and test splits used by ~\cite{oh2016deep, zheng2019hardness}.
(\romannum{1}) CUB200-2011~\cite{wah2011caltech} (CUB200) contains 200 different bird species with 11,788 images in total.
The first 100 classes with 5,864 images are used for training, and the other 100 classes with 5,924 images are used for testing.
(\romannum{2}) CARS196~\cite{krause20133d} contains 196 different types of cars with 16,185 images.
The first 98 classes with 8,054 images are used for training, and the other 98 classes with 8,131 images are used for testing.
(\romannum{3}) Standford Online Products~\cite{oh2016deep} (SOP) is one of the largest benchmarks for the metric learning task.
It consists of 22,634 classes of online products with 120,053 images, where 11,318 classes with 59,551 images are used for training, and the other 11,316 classes with 60,052 images are used for testing.
For CUB200 and CARS196, we evaluate the proposed method without bounding box information.

\paragraph{Metrics}
Following the standard metrics in image retrieval and clustering~\cite{oh2016deep, wang2017deep}, we report the image clustering performance with $F_1$ and normalized mutual information (NMI) metrics~\cite{schutze2008introduction} and image retrieval performance with Recall@K score.

\paragraph{Experimental Settings}
We implement our proposed method with the TensorFlow~\cite{abadi2016tensorflow} framework on a Tesla P40 GPU with 24GB memory.
Input images are resized to 256 $\times$ 256, horizontally flipped, and randomly cropped to 227 $\times$ 227.
We use a 512-dimensional embedding size for all feature vectors.
All models are trained with an ImageNet~\cite{deng2009imagenet} pre-trained GoogLeNet~\cite{szegedy2015going} and a randomly initialized fully connected layer using the Xavier method~\cite{glorot2010understanding}.
We use the learning rate of $10^{-4}$ with the Adam optimizer~\cite{kingma2014adam} and set a batch size of 128 for every dataset.
For the baseline metric learning loss, we use triplet loss with hard positive and hard negative mining (HPHN)~\cite{hermans2017defense, xuan2019improved} and its combination of EE with the number of synthetic points $n=2$ across all experiments, unless otherwise noted in the experiment.

\begin{figure}[t!]
    \centering
     \begin{subfigure}[b]{0.5\columnwidth}
        \centering\includegraphics[width=1.01\columnwidth]{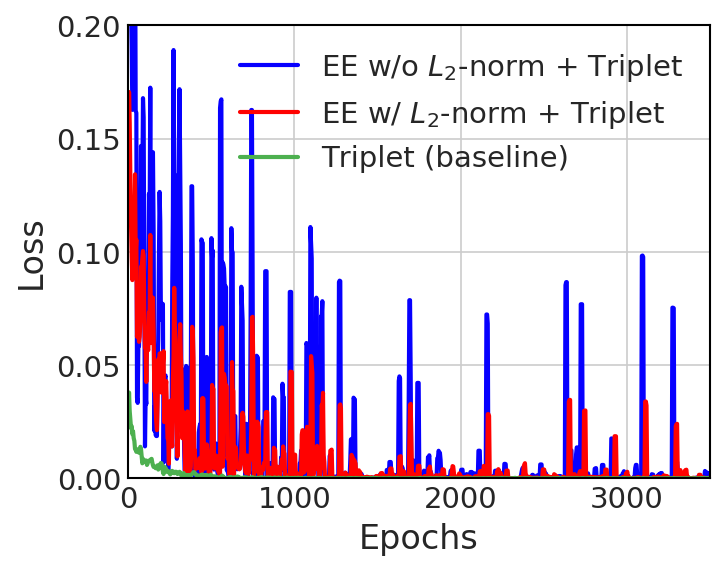}
         \caption{Loss value}
         \label{fig:y equals x}
     \end{subfigure}\hfill
    \begin{subfigure}[b]{0.5\columnwidth}
        \centering\includegraphics[width=1.0\columnwidth]{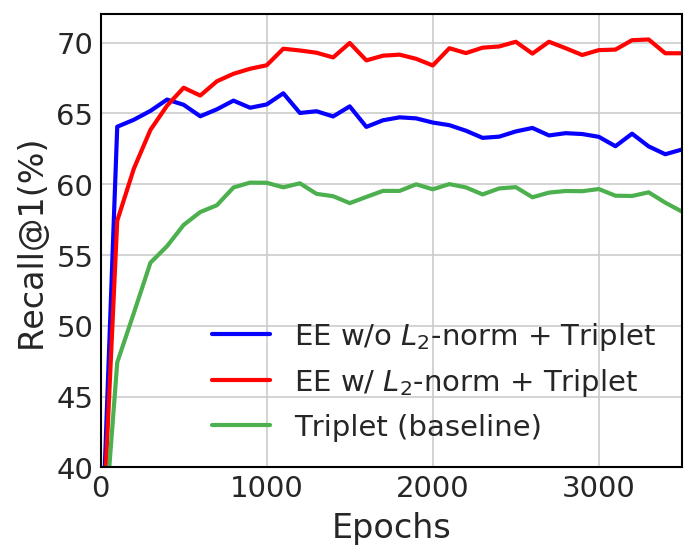}
         \caption{Recall@1(\%) performance}
         \label{fig:y equals x}
     \end{subfigure}
\caption{Loss value and recall@1(\%) performance of training and test set from CARS196.
It compares three models: triplet loss as baseline, EE without $L_2$-normalization + triplet loss, and EE with $L_2$-normalization + triplet loss.}
\label{fig:l2-norm}
\end{figure}

\subsection{Analysis of Synthetic Points} \label{sec:synthetic}

\subsubsection{Labels of Synthetic Points} \label{sec:label}

The main advantage of exploiting the internally dividing point is that the labels of synthetic points are expected to have a high degree of certainty because they are placed inside the class cluster.
Thus, they can contribute to training a network as synthetic points with augmented information other than outliers.
To investigate the certainty of the synthetic points during the training phase, we conduct an experiment that the synthetic and original points from the train set are evaluated at each epoch.
For the evaluation of the synthetic points, we used the synthetic points as the query side and the original points as the database side.
The score of synthetic points at the beginning is above 80\%, which is enough for training, and it starts increasing by the training epoch.
Overall, recall@1 of synthetic points from train sets are always higher than those of original points, and they maintain a high degree of certainty to be used as augmented feature points.

\subsubsection{Impact of $L_2$-normalization}
Metric learning losses, such as triplet, lifted structured, and MS loss, apply $L_2$-normalization to the last feature embeddings so that every feature embedding will be projected onto the hyper-sphere space with the same norm.
Generating internally dividing points between $L_2$-normalized feature embeddings will not be on the hyper-sphere space and will have a different norm.
Thus, we proposed applying $L_2$-normalization to the synthetic points to keep the continuity of the norm for these kinds of metric learning losses.
To investigate the impact of $L_2$-normalization, we conduct an experiment of EE with and without $L_2$-normalization, including a baseline of triplet loss, as illustrated in Figure~\ref{fig:l2-norm}.
Interestingly, EE without $L_2$-normalization achieved better performance than the baseline.
However, the model's loss value fluctuates greatly, which can be caused by the different norms between original and synthetic points.
The baseline and EE without $L_2$-normalization start decreasing after peak points, which indicates the models are overfitting on the training set.
On the other hand, the performance graph of EE with $L_2$-normalization keeps increasing because of training with augmented information, which enables one to obtain a more generalized model.

\begin{figure}[t!]
\centering
\includegraphics[width=0.8\columnwidth]{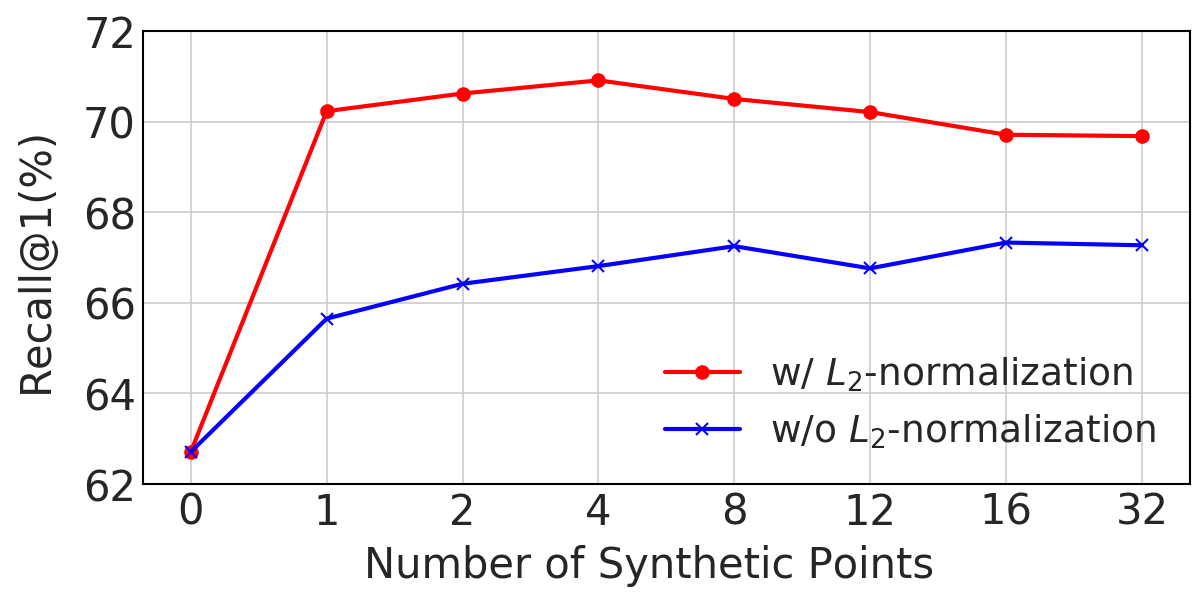} 
\caption{Recall@1(\%) performance by the number of synthetic points with and without $L_2$-normalization.
Each model is trained with EE + triplet loss on the CARS196.}
\label{fig:num_point}
\end{figure}

\subsubsection{Impact of Number of Synthetic Points}

The number of synthetic points to generate is the sole hyper-parameter of our proposed method.
As illustrated in Figure~\ref{fig:num_point}, we conduct an experiment by differentiating the number of synthetic points on EE with and without $L_2$-normalization to see its impact.
For EE without $L_2$-normalization, the performance keeps increasing until about 8 synthetic points and maintains performance.
In the case of the EE with $L_2$-normalization, the peak of the performance is between 2 and 8 synthetic points, after which it starts decreasing.
We speculate that it is because generating too many synthetic points can cause the model to be distracted by the synthetic points.

\begin{figure}[t!h!]
    \centering
    \begin{subfigure}[b]{0.5\columnwidth}
        \centering\includegraphics[width=1.0\columnwidth]{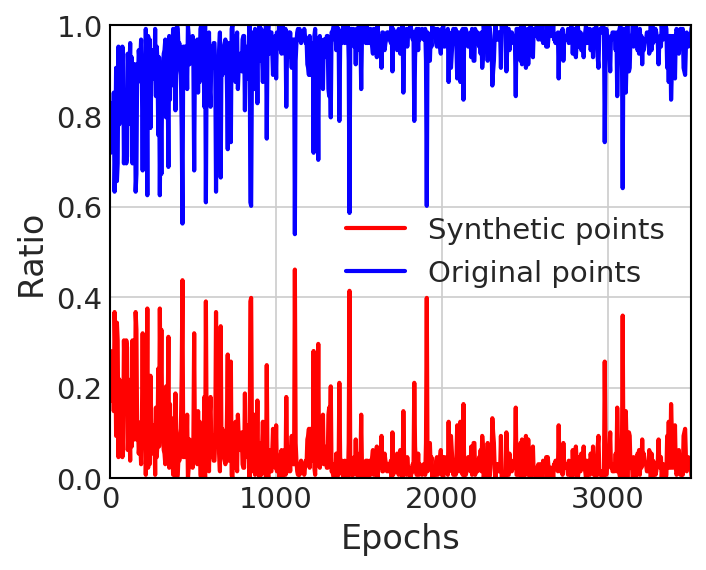}
         \caption{$n=2$}
         \label{fig:y equals x}
     \end{subfigure}\hfill
     \begin{subfigure}[b]{0.5\columnwidth}
        \centering\includegraphics[width=1.0\columnwidth]{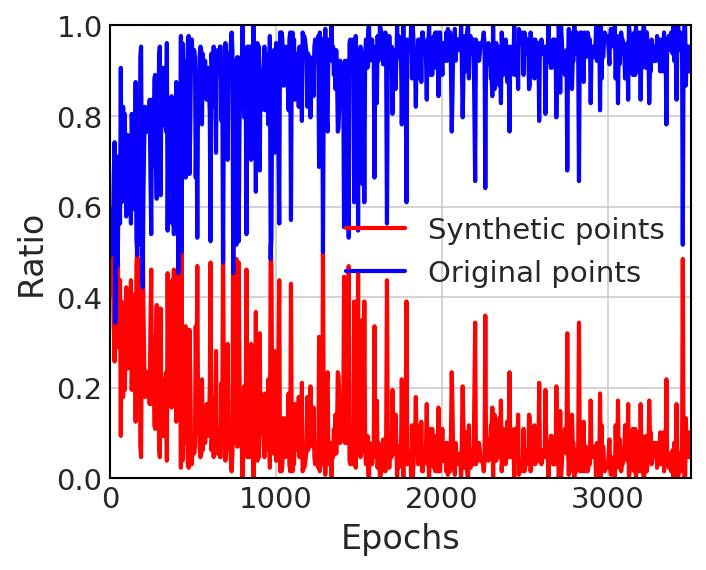}
         \caption{$n=16$}
         \label{fig:y equals x}
     \end{subfigure}
\caption{Ratio of synthetic and original points which are selected during hard negative pair mining of EE + triplet loss. We generate $n$ synthetic points for EE and train the model with CARS196. The ratio of synthetic point is calculated as $ratio^{(syn)} = \frac{\#\,of\,synthetic}{\#\,of\,synthetic\,+\,\#\,of\,original}$, while the ratio of original point is calculated as $ratio^{(ori)} = 1 - ratio^{(syn)}$.}
\label{fig:syn_ratio}
\end{figure}

\subsection{Analysis of Hard Negative Pair Mining} \label{sec:mining}

\subsubsection{Selection Ratio of Synthetic Points}

The proposed method performs hard negative pair mining among synthetic and original points in order to learn the metric learning loss with the most informative feature representations.
To see the impact of synthetic points, we compute the ratio of synthetic and original points selected in the hard negative pair mining, as illustrated in Figure~\ref{fig:syn_ratio}.
At the beginning of training, more than 20\% of synthetic points are selected for the hard negative pair.
The ratio of synthetic points decreases as the clustering ability increases because many synthetic points are generated inside of the cluster.
By increasing the number of $n$, the ratio of synthetic points increases.
Throughout the training, a greater number of original points are selected than synthetic points.
This way, the synthetic points work as assistive augmented information instead of distracting the model training.

\begin{figure*}[t!]
\centering
\includegraphics[width=0.8\textwidth]{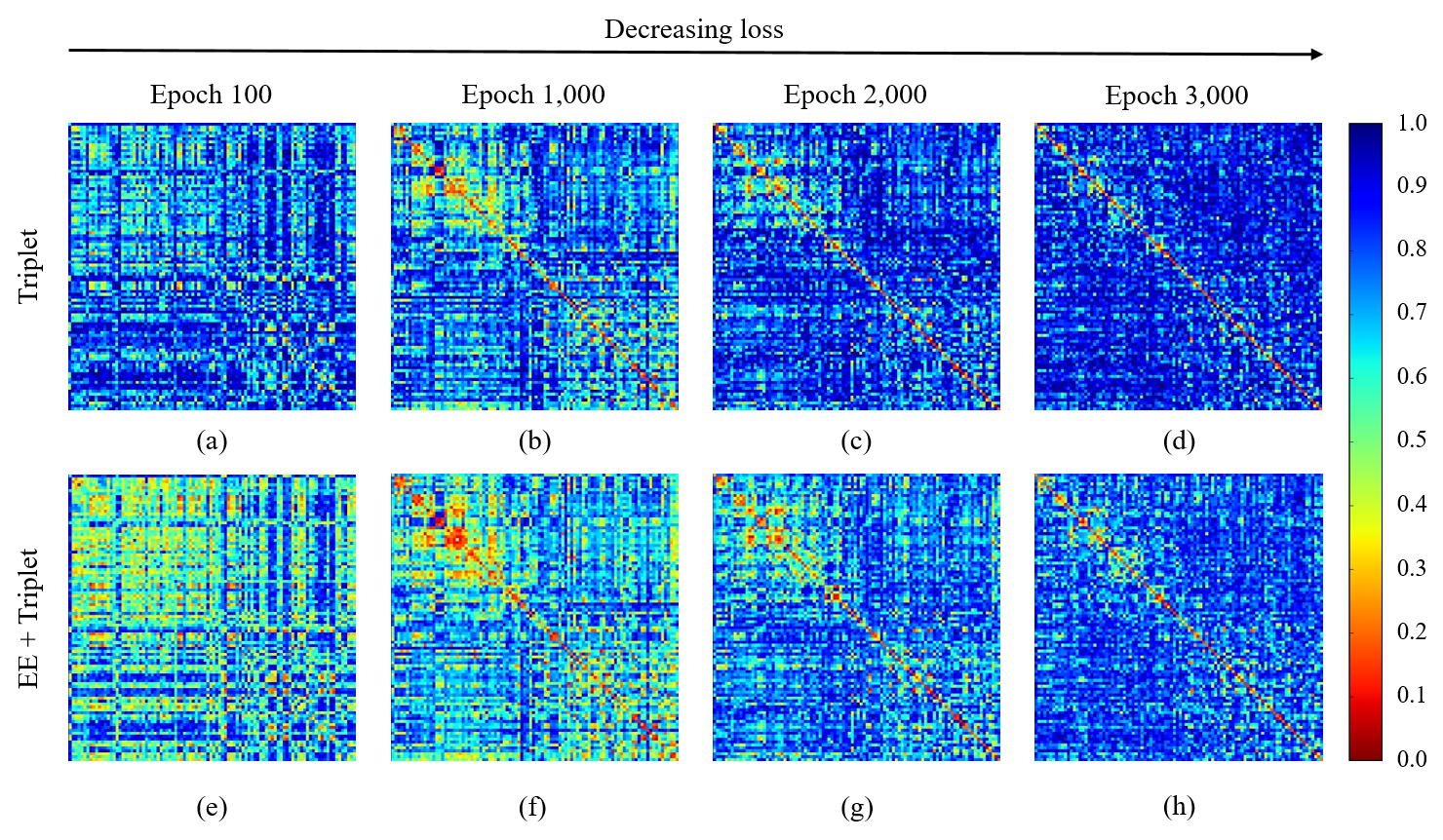} 
\caption{Comparison of Euclidean distance heatmaps between triplet and EE + triplet loss during training CARS196 dataset. In each heatmap, given two samples from each class, all the rows and columns are the first and the second samples from each class, respectively. The main diagonal is the distance of positive pairs, where the entries outside the diagonal are the distance of negative pairs. The smaller distance of negative pair ({\color{yellow}yellow} and {\color{red}red}) indicates the harder negative pairs, where all distance is normalized between 0 and 1.}
\label{fig:heatmap}
\end{figure*}

\subsubsection{Effect of Hard Negative Pair Mining}

We visualized distance matrices of triplet loss as a baseline, and EE + triplet loss to see the effect of hard negative pair mining as illustrated in Figure~\ref{fig:heatmap}.
By increasing the training epoch, the main diagonal of the heatmaps get redder, and the entries outside the diagonal get bluer in both triplet and EE + triplet loss.
This indicates that the distances of positive pairs get smaller with smaller intra-class variation, and the distances of negative pairs get larger with a larger inter-class variation.
In a comparison of triplet and EE + triplet loss on the same epoch, heatmaps of EE + triplet loss are filled with more yellow and red colors than the baseline of triplet, especially at the beginning of the training as shown in Figure~\ref{fig:heatmap}e and Figure~\ref{fig:heatmap}f.
Even at the end of the training, as Figure~\ref{fig:heatmap}h, the heatmap of EE + triplet loss still contains a greater number of hard negative pairs than triplet loss does.
It shows that combining the proposed embedding expansion with the metric learning loss allows training with harder negative pairs with augmented information.
A more detailed analysis of the training process is presented in supplementary material and video\footnote{\url{https://youtu.be/5msMSXyQZ5U}} with a t-SNE visualization~\cite{maaten2008visualizing} of embedding space at certain epoch.

\subsection{Analysis of Model} \label{sec:model}

\begin{figure}[t!]
    \centering
    \begin{subfigure}[b]{0.5\columnwidth}
        \centering\includegraphics[width=1.0\columnwidth]{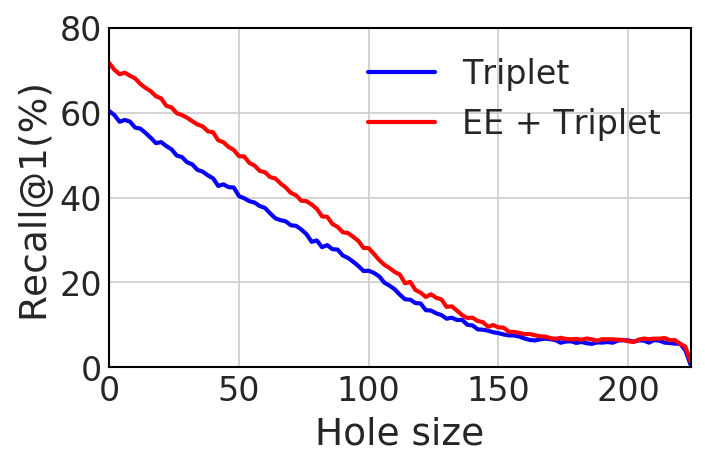}
         \caption{Center occlusion}
         \label{fig:y equals x}
     \end{subfigure}\hfill
     \begin{subfigure}[b]{0.5\columnwidth}
        \centering\includegraphics[width=1.0\columnwidth]{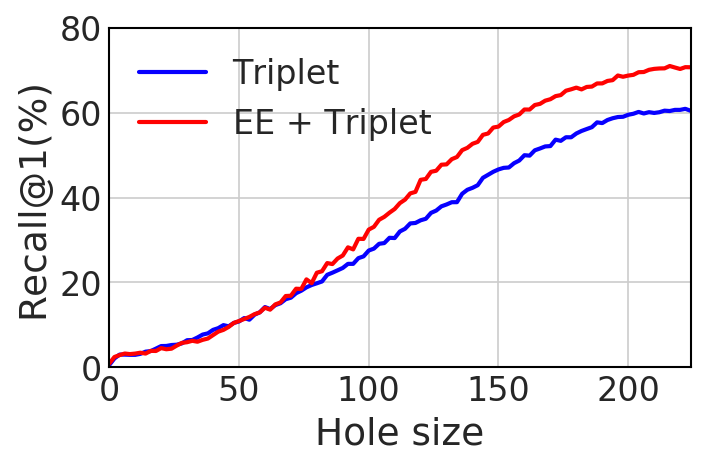}
         \caption{Boundary occlusion}
         \label{fig:y equals x}
     \end{subfigure}
\caption{Evaluation of occluded images on CARS196 test set.}
\label{fig:occlusion}
\end{figure}

\subsubsection{Robustness}
To see the improvement of the model robustness, we evaluate performance by putting occlusion on the input images in two ways.
Center occlusion fills zeros in a center hole, and boundary occlusion fills zeros outside of the hole.
As shown in Figure~\ref{fig:occlusion}, we measure Recall@1 performance by increasing the size of the hole from 0 to 227.
The result shows that EE + triplet loss achieves significant improvements in robustness for both occlusion cases.

\begin{figure}[t!]
    \centering
    \begin{subfigure}[b]{0.5\columnwidth}
        \centering\includegraphics[width=1.0\columnwidth]{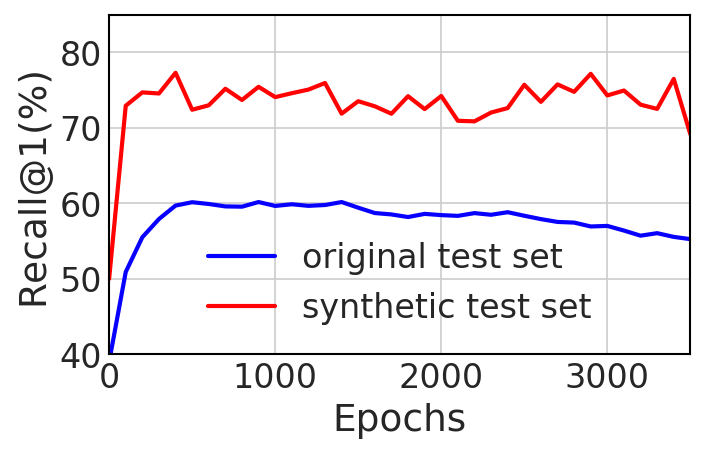}
         \caption{Triplet} 
         \label{fig:y equals x}
     \end{subfigure}\hfill
     \begin{subfigure}[b]{0.5\columnwidth}
        \centering\includegraphics[width=1.0\columnwidth]{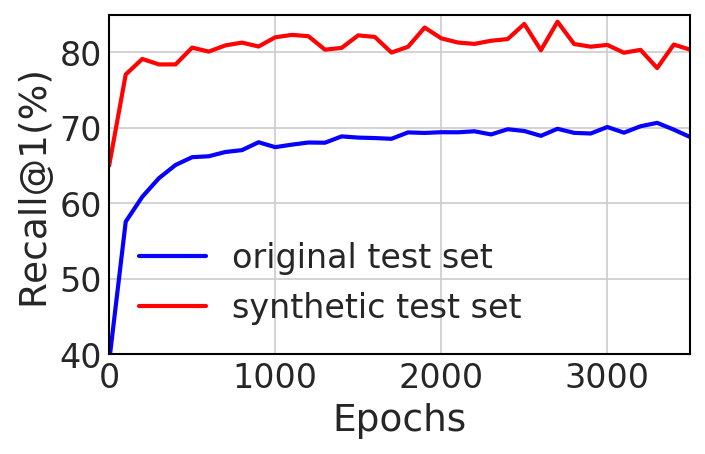}
         \caption{EE + Triplet}
         \label{fig:y equals x}
     \end{subfigure}
\caption{Evaluation of synthetic features from CARS196 test set.}
\label{fig:feature}
\end{figure}

If embedding features are well clustered and contain the key representations of the class, the combination of embedding features from the same class would be included in the same class cluster with the same effect of QE~\cite{chum2007QE, chum2011QE} and DBA~\cite{turcot2009DBA, arandjelovic2012DBA}.
To see the robustness of the embedding features, we generate synthetic points by combining randomly selected test points of the same class as a query side and evaluate Recall@1 performance with original test points as a database side.
As shown in Figure~\ref{fig:feature}, EE + triplet loss improves the performance of evaluation with synthetic test sets compared to triplet loss, which shows that the feature robustness is improved by forming well-structured clusters.

\subsubsection{Training Speed and Memory} \label{sec:speedmemory}

Additional training time and memory of the proposed method are negligible.
To see the additional training time of the proposed method, we compared the training time of baseline triplet and EE + triplet loss.
As shown in Table~\ref{table:speed}, generating synthetic points takes from 0.0002 ms to 0.0023 ms longer.
The total computing time of EE + triplet loss takes just 0.0038 ms to 0.0159 ms longer than the baseline ($n=0$).
Even though the number of points to generate increases, the additional time for computing is negligible because it can be done by simple linear algebra.
For memory requirements, if $N$ embedding points with a $N \times N$ similarity matrix are necessary for computing the triplet loss, EE + triplet loss requires $(n+1)N$ embedding points with a $(n+1)N \times (n+1)N$ similarity matrix, which are trivial.

\begin{table}[t!]
\begin{center}
\begin{adjustbox}{width=1.0\columnwidth,center}
\begin{tabular}{c|cccccc}
\Xhline{2\arrayrulewidth}
$n$   & 0  & 2  & 4  & 8  & 16 & 32   \\ \hline\hline
Gen   & 0      & 0.0002 & 0.0004 & 0.0008 & 0.0013 & 0.0023 \\ 
Total & 0.2734 & 0.2772 & 0.2775 & 0.2822 & 0.2885 & 0.2893 \\ 
\Xhline{2\arrayrulewidth}
\end{tabular}
\end{adjustbox}
\end{center}
\caption{Computation time (ms) of EE by the number of synthetic points $n$. Gen is time for generating synthetic points, and total is time for computing triplet loss, including generation and hard negative pair mining.}
\label{table:speed}
\end{table}

\begin{table}[h!]
\begin{center}
\begin{subtable}[h!]{0.9\columnwidth}
\begin{adjustbox}{width=1.0\columnwidth,center}
\begin{tabular}{c|cc|cccc}
\Xhline{2\arrayrulewidth}
\multirow{2}{*}{Method} & \multicolumn{2}{c|}{Clustering} & \multicolumn{4}{c}{Retrieval} \\ \cline{2-7} 
                        & NMI            & F1             & R@1   & R@2   & R@4   & R@8   \\ \hline\hline
Triplet                 & 49.8           & 15.0           & 35.9  & 47.7  & 59.1  & 70.0  \\
Triplet$^\dagger$            & 58.1           & 24.2           & 48.3  & 61.9  & 73.0  & 82.3  \\
DAML (Triplet)          & 51.3           & 17.6           & 37.6  & 49.3  & 61.3  & 74.4  \\
HDML (Triplet)          & 55.1           & 21.9           & 43.6  & 55.8  & 67.7  & 78.3  \\
EE + Triplet            & 55.7           & 22.4           & 44.3  & 57.0  & 68.1  & 78.9  \\
EE + Triplet$^\dagger$  & \textbf{60.5} & \textbf{27.0} & \textbf{51.7} & \textbf{63.5} & \textbf{74.5} & \textbf{82.5}  \\ \hline
N-pair                  & 60.2           & 28.2           & 51.9  & 64.3  & 74.9  & 83.2  \\
DAML (N-pair)           & 61.3           & 29.5           & 52.7  & 65.4  & 75.5  & 84.3  \\
HDML (N-pair)           & 62.6           & 31.6  & 53.7  & 65.7  & 76.7  & 85.7  \\
EE + N-pair             & \textbf{62.7}  & \textbf{32.4} & \textbf{55.2} & \textbf{67.4} & \textbf{77.7} & \textbf{86.4}  \\ \hline
Lifted-Struct           & 56.4           & 22.6           & 46.9  & 59.8  & 71.2  & 81.5  \\
EE + Lifted             & \textbf{61.2}  & \textbf{28.2}  & \textbf{54.2} & \textbf{66.6} & \textbf{76.7} & \textbf{85.2}  \\ \hline
MS                 & 62.8  & 31.2  & 56.2  & 68.3  & 79.1  & 86.5  \\
EE + MS            & \textbf{63.3} & \textbf{32.5} & \textbf{57.4} & \textbf{68.7} & \textbf{79.5} & \textbf{86.9}  \\
\Xhline{2\arrayrulewidth}
\end{tabular}
\end{adjustbox}
\caption{CUB200-2011 dataset}
\end{subtable}
\end{center}
\vspace{-2mm}
\begin{center}
\begin{subtable}[h!]{0.9\columnwidth}
\begin{adjustbox}{width=1.0\columnwidth,center}
\begin{tabular}{c|cc|cccc}
\Xhline{2\arrayrulewidth}
\multirow{2}{*}{Method} & \multicolumn{2}{c|}{Clustering} & \multicolumn{4}{c}{Retrieval} \\ \cline{2-7} 
                        & NMI            & F1             & R@1   & R@2   & R@4   & R@8   \\ \hline\hline
Triplet                 & 52.9           & 17.9           & 45.1  & 57.4  & 69.7  & 79.2  \\
Triplet$^\dagger$       & 57.4           & 22.6           & 60.3  & 73.4  & 83.5  & 90.5  \\
DAML (Triplet)          & 56.5           & 22.9           & 60.6  & 72.5  & 82.5  & 89.9  \\
HDML (Triplet)          & 59.4           & 27.2           & 61.0  & 72.6  & 80.7  & 88.5  \\
EE + Triplet            & 60.3         & 25.1  & 57.2  & 70.5  & 81.3  & 88.2  \\
EE + Triplet$^\dagger$  & \textbf{63.1}  & \textbf{32.0}   & \textbf{71.6}  & \textbf{80.7}  & \textbf{87.5}  & \textbf{92.2}  \\ \hline
N-pair                  & 62.7           & 31.8           & 68.9  & 78.9  & 85.8  & 90.9  \\
DAML (N-pair)           & 66.0           & 36.4           & 75.1  & 83.8  & 89.7  & 93.5  \\
HDML (N-pair)           & \textbf{69.7}  & \textbf{41.6}  & \textbf{79.1}  & \textbf{87.1}  & \textbf{92.1}  & \textbf{95.5}  \\
EE + N-pair             & 63.4           & 32.6           & 72.5  & 81.1  & 87.6  & 92.5  \\ \hline
Lifted-Struct           & 57.8           & 25.1           & 59.9  & 70.4  & 79.6  & 87.0  \\
EE + Lifted             & \textbf{59.1}  & \textbf{27.2}  & \textbf{65.2}  & \textbf{76.4}  & \textbf{85.6}  & \textbf{89.5}  \\ \hline
MS                 & 62.4           & 30.2           & 75.0  & 83.1  & 89.5  & 93.6  \\
EE + MS            & \textbf{63.5}  & \textbf{33.5}  & \textbf{76.1}  & \textbf{84.2}  & \textbf{89.8}  & \textbf{93.8}  \\
\Xhline{2\arrayrulewidth}
\end{tabular}
\end{adjustbox}
\caption{CARS196 dataset}
\end{subtable}
\end{center}
\vspace{-2mm}
\begin{center}
\begin{subtable}[h!]{0.9\columnwidth}
\begin{adjustbox}{width=0.95\columnwidth,center}
\begin{tabular}{c|cc|ccc}
\Xhline{2\arrayrulewidth}
\multirow{2}{*}{Method} & \multicolumn{2}{c|}{Clustering} & \multicolumn{3}{c}{Retrieval} \\ \cline{2-6} 
                        & NMI            & F1             & R@1      & R@10     & R@100    \\ \hline\hline
Triplet                 & 86.3           & 20.2           & 53.9     & 72.1     & 85.7     \\
Triplet$^\dagger$       & 91.4           & 43.3           & 75.5     & 88.8     & 95.4     \\
DAML (Triplet)          & 87.1           & 22.3           & 58.1     & 75.0     & 88.0     \\
HDML (Triplet)          & 87.2           & 22.5           & 58.5     & 75.5     & 88.3     \\
EE + Triplet            & 87.4           & 24.8            & 62.4     & 79.0      & 91.0      \\
EE + Triplet$^\dagger$  & \textbf{91.5}  & \textbf{43.6}  & \textbf{77.2} & \textbf{89.6} & \textbf{95.5} \\ \hline
N-pair                  & 87.9           & 27.1           & 66.4     & 82.9     & 92.1     \\
DAML (N-pair)           & 89.4           & 32.4           & 68.4     & 83.5     & 92.3     \\
HDML (N-pair)           & 89.3           & 32.2           & 68.7     & 83.2     & 92.4     \\
EE + N-pair             & \textbf{90.6}  & \textbf{38.8}  & \textbf{73.5} & \textbf{87.5} & \textbf{94.4} \\ \hline
Lifted-Struct           & 87.2           & 25.3           & 62.6     & 80.9     & 91.2     \\
EE + Lifted             & \textbf{89.6}  & \textbf{35.3}  & \textbf{70.6} & \textbf{85.5} & \textbf{93.6} \\ \hline
MS                 & 91.3           & 42.7           & 75.9     & 89.3     & 95.6     \\
EE + MS            & \textbf{91.9}  & \textbf{46.1}  & \textbf{78.1} & \textbf{90.3} & \textbf{95.8} \\
\Xhline{2\arrayrulewidth}
\end{tabular}
\end{adjustbox}
\caption{Stanford Online Products dataset}
\end{subtable}
\end{center}

\caption{Clustering and retrieval performance (\%) on three benchmarks in comparison with other methods. $^\dagger$ denotes the HPHN triplet loss, and bold numbers indicate the best score within the same loss.}
\label{table:sota}
\end{table}

\subsection{Comparison with State-of-the-Art} \label{sec:sota}

We compare the performance of our proposed method with the state-of-the-art metric learning losses and previous sample generation methods on clustering and image retrieval tasks.
As shown in Table~\ref{table:sota}, all combinations of EE with metric learning losses achieve significant performance boost on both tasks. 
The maximum improvements of Recall@1 and NMI are 12.1\% and 7.4\% from triplet loss in the CARS196 dataset, respectively.
The minimum improvements of Recall@1 and NMI are 1.1\% from MS loss in the CARS196 dataset and 0.1\% from HPHN triplet loss in the SOP dataset, respectively.
By comparison with previous sample generation methods, the proposed method outperforms every dataset and loss, except for the N-pair loss on the CARS196 dataset.
In the large-scale datasets with enormous categories like SOP, the performance improvement of the proposed method is more competitive, even when taking those methods with generative networks into consideration.
Performance with different capacity of networks and comparison with HTG are presented in the supplementary material.

\section{Conclusion}
In this paper, we proposed embedding expansion for augmentation in the embedding space that can be used with existing pair-based metric learning losses.
We do so by generating synthetic points within positive pairs, and performing hard negative pair mining among synthetic and original points.
Embedding expansion is simple, easy to implement, no computational overheads, but adjustable on various pair-based metric learning losses.
We demonstrated that the proposed method significantly improves the performance of existing metric learning losses, and also outperforms the previous sample generation methods in both image retrieval and clustering tasks.

\paragraph{Acknowledgement}
We would like to thank Hyong-Keun Kook, Minchul Shin, Sanghyuk Park, and Tae Kwan Lee from Naver Clova vision team for valuable comments and discussion.

{\small
\bibliographystyle{ieee_fullname}
\bibliography{main}
}

\clearpage



\section*{Supplementary material}

# Embedding Expansion: Augmentation in Embedding Space for Deep Metric Learning

*Supplementary Material*

## A. Introduction

This supplementary material provides more details of the proposed embedding expansion (EE). First, we compare the proposed method with mixup augmentation techniques and compare the generation methods between EE and mixup. Then, we show the visualization of the embedding space to see the effect of the proposed method. Finally, we investigate the impact of network capacity to see if the proposed method works for different sizes of models.

## B. Comparison with MixUp

Early works of mixup [10, 7] propose a data augmentation method by combining two input samples, where the ground truth label of the combined sample is given by the mixture of one-hot labels. By doing so, it improves the generalization of the neural network by regularizing the network to behave linearly in-between training samples. While those mixup methods work in the input space, manifold mixup [8] performs linear combinations of hidden representations of training samples in the representation space. It also improves the generalization of the neural network by perturbing the hidden representations, similar to dropout [6], batch normalization [4], and information bottleneck [1]. In the representative input mixup [10, 7], generating virtual feature-target vectors $(\tilde{x}, \tilde{y})$ is formulated as:

$$\tilde{x} = \lambda x_i + (1 - \lambda)x_j, \quad \text{(i)}$$
$$\tilde{y} = \lambda y_i + (1 - \lambda)y_j, \quad \text{(ii)}$$

where $(x_i, y_i)$ and $(x_j, y_j)$ are two feature-target vectors from the training data, $\lambda \sim Beta(\alpha, \alpha)$ for $\alpha \in (0, \infty)$, and $\lambda \in [0, 1]$.

The proposed embedding expansion has similarity with the mixup techniques, where both methods generate virtual feature vectors by combining two original feature vectors for augmentation. However, both methods have major differences in four points: (i) The proposed embedding expansion is for pair-based metric learning losses, whereas the mixup is for softmax loss and its variants. The mixup can not be used with pair-based metric learning losses because most of the pair-based metric learning losses require obvious class labels and can not exploit the mixture of one-

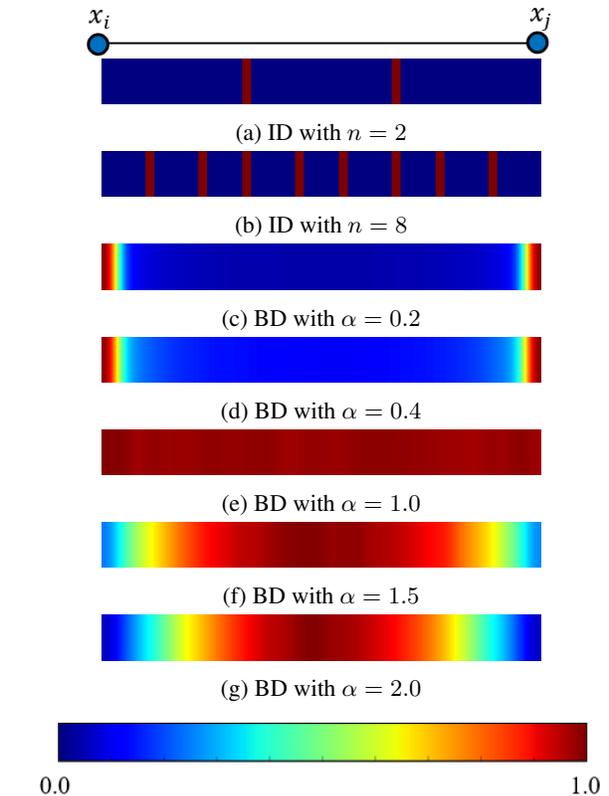

Figure A. A visualization of the locational generation ratio between an original pair ($x_i$ and $x_j$) from the same class with two different generation methods: the proposed internally dividing (ID) points into $n + 1$ equal parts and beta distribution (BD) with $\alpha$ parameter.

hot labels. (ii) The proposed method generates synthetic points in-between a pair from the same class, which are internally dividing points into $n + 1$ equal parts, while the mixup uses mixing coefficient $\lambda$ sampling from beta distribution and generates virtual feature vectors in-between a pair from the different class. (iii) The proposed method exploits the output embedding feature points from a network, while the mixup techniques use feature vectors from the input or hidden representation of a network. (iv) After gener-

i

ating synthetic points, the proposed method performs hard negative pair mining to select the most informative feature points among original and synthetic points.

### B.1. Generation with Beta Distribution

The proposed EE generates synthetic points between a pair, which are internally dividing points into $n+1$ equal parts. The synthetic points will be generated on the deterministic locations, as illustrated in Figure Aa and Ab. On the other hand, it is possible to generate synthetic points by using the beta distribution as Equation i of mixup. This will generates synthetic points on the stochastic locations. Smaller $\alpha$ values generates synthetic points nearby the original points (Figure Ac and Ad) and larger $\alpha$ values generates synthetic points around middle of the pair (Figure Af and Ag), while $\alpha = 1.0$ is equal to the uniform distribution (Figure Ae).

To compare these two generation methods, we conduct an experiment by generating $n$ synthetic points with these generation methods and use the same hard negative pair mining as proposed EE. We use triplet loss with hard positive and hard negative (HPHN) mining [3, 9], trained with CARS196 dataset. As shown in Table A, methods of EE (BD) with $\alpha = 0.2$ outperform the baseline model, while larger $\alpha$ decreases the performance. It indicates that the stochastic generation between a pair can be distractive for the training, except for the synthetic points which are close and similar to the original points. Meanwhile, the proposed EE (ID) shows better performance than the baseline and the EE (BD), which shows that the deterministic generation is more stable and effective.

### C. Visualization of Embedding Space

In order to see the process of clustering during training, we visualize the embedding space of certain training epochs with the Barnes-Hut t-SNE [5]. We use HPHN triplet loss in Figure B and its combination of EE in Figure C, trained with CARS196 dataset. For each model, we visualize the embedding of the train data with different colors for different classes, and the joint embedding of the train and test data to highlight where the test data is embedded compared to train data.

At the beginning of the training, the train and the test set of both triplet and EE + triplet are scattered without forming any discriminative clusters (Figure Ba, Bd, Ca, and Cd). In the middle of the training at 1000 epoch, the train set of EE + triplet starts having clusters (Figure Cb) and the test set are less scattered than the 10 epoch (Figure Ce), while the train set of triplet also starts having clusters with less inter-class variation than EE + triplet (Figure Bb). At the end of the training at 3000 epoch, the train set of EE + triplet has more discriminative clusters with larger inter-class variation, compared to the triplet embedding (Figure Bc and Cc).

| Method | $n$ | $\alpha$ | R@1 |
|---|---|---|---|
| Baseline | 0 | - | 60.3 |
| EE (ID) | 2 | - | **71.6** |
| EE (BD) | 2 | 0.2 | 66.7 |
|  |  | 0.4 | 59.6 |
|  |  | 1.0 | 56.5 |
|  |  | 1.5 | 55.1 |
|  |  | 2.0 | 53.8 |
| EE (ID) | 8 | - | **71.2** |
| EE (BD) | 8 | 0.2 | 61.5 |
|  |  | 0.4 | 58.1 |
|  |  | 1.0 | 54.3 |
|  |  | 1.5 | 54.5 |
|  |  | 2.0 | 53.9 |

Table A. Performance (%) comparison among baseline, EE (ID), and EE (BD) with HPHN triplet trained on CARS196. We generate $n$ synthetic points by using the proposed internally dividing points into $n+1$ equal parts for EE (ID) and beta distribution with $\alpha$ parameter for EE (BD).

The test set of EE + triplet are less spread out and forming some clusters compared to the triplet embedding (Figure Cf and Bf). Overall, the combination of triplet loss and the proposed method has shown a better clustering ability than the sole triplet loss. Entire visualization of the training process can be found in the supplementary video.

### D. Impact of Network Capacity

In order to see the impact of network capacity on the proposed method, we conduct an experiment by differentiating the network capacity and the number of synthetic points. We used one of the most generally used ResNet50_v1 [2] and its smaller capacity variants (ResNet18_v1 and ResNet34_v1). Moreover, we compare the proposed method with the hard triplet generation (HTG) [11] method on the same network capacity of ResNet18_v1. Throughout the experiment, we use HPHN triplet and its combination with the proposed method on CUB200-2011 (CUB200), CARS196, and stanford online products (SOP) datasets.

As shown in the Table B, the proposed method achieves around 1% to 3% of performance boost for every network and dataset. We observe that there are the best number of synthetic points $n$ for each dataset, such as $n = 8$ for CUB200, $n = 4$ for CARS196, and $n = 2$ for SOP. In comparison with HTG, even though it uses a combination of no-bias softmax and triplet loss, and a generative adversarial network for sample generation, the proposed method outperforms for every dataset.

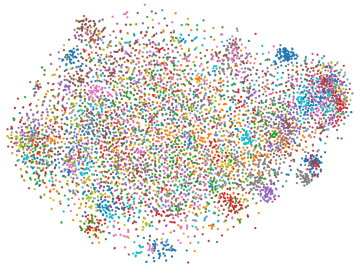
(a) Train, 10 epoch

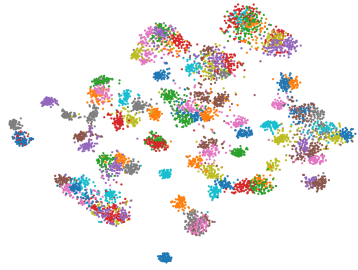
(b) Train, 1000 epoch

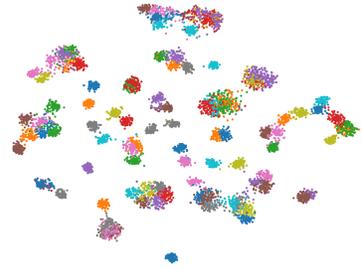
(c) Train, 3000 epoch

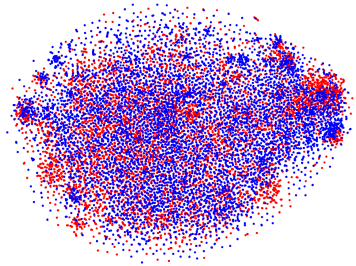
(d) Train + test, 10 epoch

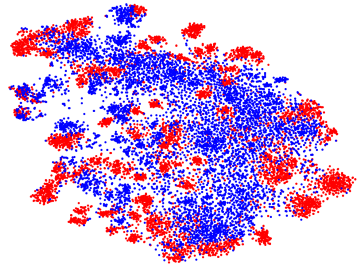
(e) Train + test, 1000 epoch

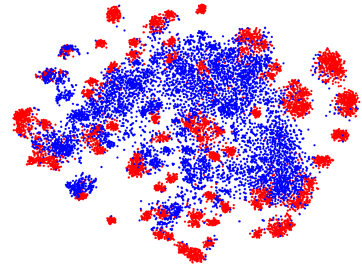
(f) Train + test, 3000 epoch

Figure B. A t-SNE visualization of triplet loss with CARS196 dataset. (a), (b), and (c) are the embedding of the train data, while (d), (e), and (f) are the joint embedding of the train (red) and the test (blue) data at each epoch.

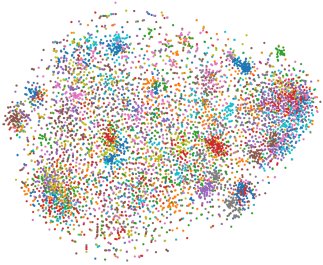
(a) Train, 10 epoch

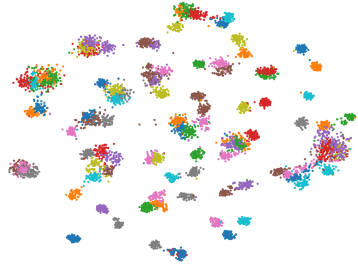
(b) Train, 1000 epoch

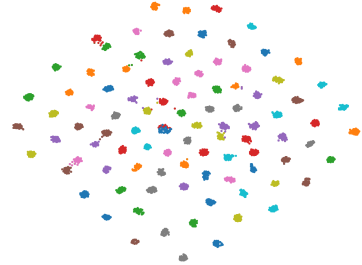
(c) Train, 3000 epoch

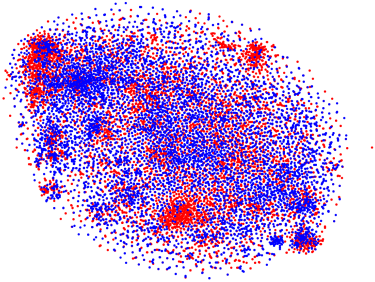
(d) Train + test, 10 epoch

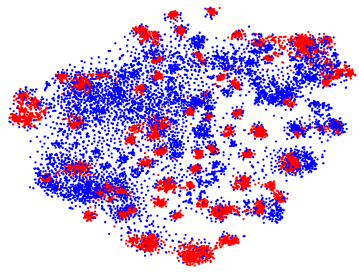
(e) Train + test, 1000 epoch

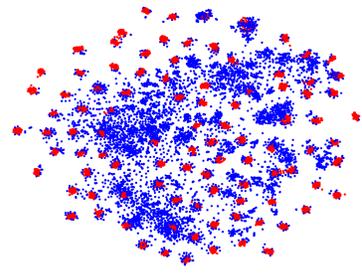
(f) Train + test, 3000 epoch

Figure C. A t-SNE visualization of EE + triplet loss with CARS196 dataset. (a), (b), and (c) are the embedding of the train data, while (d), (e), and (f) are the joint embedding of the train (red) and the test (blue) data at each epoch.

| Method | Network | n | CUB200 | | | | CARS196 | | | | SOP | | |
|---|---|---|---|---|---|---|---|---|---|---|---|---|---|
| | | | R@1 | R@2 | R@4 | R@8 | R@1 | R@2 | R@4 | R@8 | R@1 | R@10 | R@100 |
| HTG (Soft.+Tri.)[†] | ResNet18_v1[*] | - | 59.5 | 71.8 | 81.3 | 88.2 | 76.5 | 84.7 | 90.4 | 94 | - | - | - |
| EE + Triplet[‡] | ResNet18_v1 | 0 | 58.6 | 70.7 | 80.6 | 88.1 | 83.5 | 90.3 | 94.6 | 97.1 | 74.7 | 88.5 | 95.2 |
| | | 2 | 59.3 | 70.7 | 80.8 | 88.1 | 84.5 | 90.8 | 94.5 | 97.0 | **76.9** | **89.4** | **95.3** |
| | | 4 | 59.6 | 70.0 | 79.8 | 87.4 | 84.5 | 90.8 | 94.7 | 97.1 | 76.8 | 89.2 | 95.2 |
| | | 8 | **60.2** | **71.9** | **81.5** | **88.4** | **85.4** | **91.4** | **94.9** | **97.2** | 76.1 | 88.8 | 94.9 |
| | ResNet34_v1 | 0 | 60.3 | 72.9 | 82.4 | 89.3 | 84.9 | 90.1 | 94.6 | 97.0 | 75.6 | 89.1 | 95.5 |
| | | 2 | 61.1 | 73.3 | 83.0 | 89.0 | 85.0 | 91.2 | 95.0 | 97.1 | **78.2** | **90.3** | **95.8** |
| | | 4 | 61.9 | 73.7 | 83.2 | 89.4 | **85.5** | **91.6** | **95.4** | **97.3** | 77.8 | 89.7 | 95.5 |
| | | 8 | **62.7** | **74.7** | **83.9** | **89.8** | 85.1 | 91.1 | 94.5 | 96.9 | 77.9 | 90.0 | 95.7 |
| | ResNet50_v1 | 0 | 63.0 | 73.9 | 83.1 | 89.4 | 87.3 | 93.0 | 96.1 | 98.0 | 81.2 | 92.0 | 96.6 |
| | | 2 | 63.5 | 74.9 | 84.3 | 89.7 | 88.2 | 93.2 | 96.2 | 97.9 | **82.5** | **92.7** | **96.9** |
| | | 4 | 63.7 | 75.0 | 83.5 | 89.8 | **88.4** | **93.6** | **96.3** | **98.1** | 82.3 | 92.4 | 96.8 |
| | | 8 | **64.6** | **75.4** | **83.9** | **90.9** | 88.3 | 93.3 | 96.1 | 98.0 | 82.0 | 92.4 | 96.8 |

Table B. Retrieval performance (%) of different network capacity, where $n = 0$ are baseline models. [†] denotes HTG method with no-bias softmax loss and triplet loss, [‡] denotes the HPHN triplet, and [*] is specifically modified ResNet18_v1.



\end{document}